\documentclass[sigconf]{acmart}

\usepackage{hyperref}
\usepackage{CJKutf8}
\usepackage{booktabs}
\usepackage{graphicx}
\usepackage{array}
\usepackage{bm}
\usepackage{amsmath}

\usepackage{balance}
\usepackage{threeparttable}
\usepackage{multirow}
\usepackage{makecell}

\usepackage{latexsym}
\newcommand{\tabincell}[2]{\begin{tabular}{@{}#1@{}}#2\end{tabular}}
\usepackage[T1]{fontenc}

\usepackage[utf8]{inputenc}
\usepackage{float}

\usepackage{microtype}

\usepackage{xcolor}

\def\BibTeX{{\rm B\kern-.05em{\sc i\kern-.025em b}\kern-.08em
    T\kern-.1667em\lower.7ex\hbox{E}\kern-.125emX}}
\usepackage{cite}
\usepackage{amsmath,amssymb,amsfonts}
\usepackage{algorithmic}
\usepackage[ruled,linesnumbered,noline,noend]{algorithm2e}

\SetCommentSty{mycommfont}

\usepackage[medium,compact]{titlesec}

\SetNlSty{}{}{}

\let\oldnl\nl
\newcommand\nonl{%
  \renewcommand{\nl}{\let\nl\oldnl}}

\hypersetup{
    colorlinks=true,
    linkcolor=blue,}
\AtBeginDocument{%
  \providecommand\BibTeX{{%
    \normalfont B\kern-0.5em{\scshape i\kern-0.25em b}\kern-0.8em\TeX}}}

\copyrightyear{2023}
\acmYear{2023}
\setcopyright{acmcopyright}\acmConference[WSDM '23]{Proceedings of the Sixteenth ACM International Conference on Web Search and Data Mining}{February 27-March 3, 2023}{Singapore, Singapore}
\acmBooktitle{Proceedings of the Sixteenth ACM International Conference on Web Search and Data Mining (WSDM '23), February 27-March 3, 2023, Singapore, Singapore}
\acmPrice{15.00}
\acmDOI{10.1145/3539597.3570431}
\acmISBN{978-1-4503-9407-9/23/02}

\begin{document}

\title{Can Pre-trained Language Models Understand Chinese Humor?}


\author{Yuyan Chen}
\email{chenyuyan21@m.fudan.edu.cn}
\orcid{0000-0002-4381-486X}
\affiliation{%
  \institution{Shanghai Key Laboratory of Data Science, School of Computer Science, Fudan University}
  \city{Shanghai}
  \country{China}
}

\author{Zhixu Li}
\authornote{The corresponding authors.}
\email{zhixuli@fudan.edu.cn}
\orcid{0000-0003-2355-288X}
\affiliation{%
  \institution{Shanghai Key Laboratory of Data Science, School of Computer Science, Fudan University}
  \city{Shanghai}
  \country{China}
}

\author{Jiaqing Liang}
\email{liangjiaqing@fudan.edu.cn}
\orcid{0000-0003-0670-5602}
\affiliation{%
  \institution{School of Data Science, Fudan University}
  \city{Shanghai}
  \country{China}
}

\author{Yanghua Xiao}
\authornotemark[1]
\email{shawyh@fudan.edu.cn}
\orcid{0000-0001-8403-9591}
\affiliation{%
  \institution{Shanghai Key Laboratory of Data Science, School of Computer Science, Fudan University \& Fudan-Aishu Cognitive Intelligence Joint Research Center}
  \city{Shanghai}
  \country{China}
}

\author{Bang Liu}
\email{bang.liu@umontreal.ca}
\orcid{0000-0002-9483-8984}
\affiliation{%
  \institution{RALI \& Mila, Université de Montréal}
  \city{Montréal}
  \state{Québec}
  \country{Canada}
}

\author{Yunwen Chen}
\email{chenyunwen@datagrand.com}
\orcid{0000-0003-4513-9439}
\affiliation{%
  \institution{DataGrand Inc.}
  \city{Shanghai}
  \country{China}
}

\renewcommand{\shortauthors}{Yuyan Chen et al.}
\begin{abstract}
Humor understanding is an important and challenging research in natural language processing. 
As the popularity of pre-trained language models (PLMs), some recent work makes preliminary attempts to adopt PLMs for humor recognition and generation.
However, these simple attempts do not substantially answer the question: {\em whether PLMs are capable of humor understanding?}
This paper is the first work that systematically investigates the humor understanding ability of PLMs.
For this purpose, a comprehensive framework with three evaluation steps and four evaluation tasks is designed.
We also construct a comprehensive Chinese humor dataset, which can fully meet all the data requirements of the proposed evaluation framework.
Our empirical study on the Chinese humor dataset yields some valuable observations, which are of great guiding value for future optimization of PLMs in humor understanding and generation.
\end{abstract}

%
%

\ccsdesc[500]{Computing methodologies~Natural language processing}

\keywords{Humor evaluation framework, Humor understanding, Chinese humor dataset, Pre-trained language models}


\maketitle

\vspace{-0.5em}
\section{INTRODUCTION}
\vspace{-0.5em}
Humor is an advanced language art prevalently used in human languages. However, it is very challenging to let machines possess a sense of humor as humans, since it requires a deep understanding of semantics as well as cultural background.
Nowadays, as the development of human-machine interaction systems and applications, how to let machines have a sense of humor has become an increasingly important topic in Natural Language Processing (NLP). Its success or failure may potentially forecast whether a Babel of human-machine interaction could finally be built.

Due to its importance, great efforts has been made on humor-relevant tasks in the NLP community, which mainly focuses on Humor Recognition and Humor Generation. 
Early work mainly relies on shallow linguistic features and templates to recognize or generate humors. For instance, \citet{Recognizing_Humour_using} and \citet{Humor_Recognition_and} recognize humor with words associations and the latent semantic structures, while \citet{Automatic_generation_of} and \citet{Pun_Generation_with} generate poetic three liner jokes and puns through analyzing the structure and retrieve-and-edit approach, respectively. 
However, these methods rely on unaffordable human cost to design features or templates for different datasets, which can only recognize or generate a very limited range of humorous expressions.

As the popularity of pre-trained language models (PLMs), some recent work, such as \citet{A_Neural_Approach} and \citet{RoMa_at_HAHA-2021}, make preliminary attempts to finetune PLMs for humor recognition and generation.
Thanks to the powerful understanding and generation capabilities that PLMs have learned from massive amounts of data, they significantly reduce human cost and enable the recognition (or generation) on more types of humorous expressions.
However, these simple endeavors do not substantially answer an important question: {\em whether PLMs are capable of humor understanding?}

This is a deep question worth exploring, which should be answered firstly when we utilize PLMs for various humor understanding and generation tasks.
To answer this question, we would like to investigate the humor understanding ability of PLMs in the following several aspects:
%
%
%
i) Whether PLMs can understand humor before or after fine-tuning?
ii) Whether existing external knowledge can help improve PLMs' humor understanding ability?
iii) Whether PLMs can detect interpretable clue words that fit human intuitive understanding of humor?
%
To this end, we need a well-designed evaluation framework and corresponding comprehensive dataset, both of which cannot be directly obtained from the existing humor-relevant tasks ~\citep{Recognizing_Humour_using,Humor_Recognition_and,Automatic_generation_of,Pun_Generation_with}.

In this paper, we first propose a three-step evaluation framework, each step of which is responsible for answering one of the above questions. Next, within this framework, we employ four representative humor-relevant tasks to conduct the systematically evaluation on PLMs, including humor recognition, humor type classification, humor level classification and punchline detection.
Meanwhile, we construct a comprehensive Chinese humor dataset, which fully meets all data requirements of the four tasks and three steps evaluation framework. We choose to construct the Chinese humor dataset since Chinese humor is as worthy of study as English humor and more challenging. However, the existing Chinese humor datasets\textsuperscript{\ref{bisai},\ref{kaiyuan}} are far less abundant than the English humor datasets, thus we want to fill in the gap for Chinese humor research.
%
%

Our empirical study based on this Chinese humor dataset suggests that:
1) By fine-tuning on the constructed humor dataset, the humor understanding ability of PLMs has been greatly improved.
%
2) Some external knowledge, such as Chinese pinyin information, has a positive effect on improving the PLMs' performance on humor-related tasks.
%
%
3) Moreover, a portion of the detected clue words are considered being in line with human perception of humor, but there is much room for improvement in PLMs' ability to understand humor.
To summarize, our contributions in this paper are threefold:
\vspace{-5pt}
\begin{itemize}
\setlength{\itemsep}{0pt}
    \item We are the first work to systematically evaluate the ability of PLMs in understanding Chinese humor. For this purpose, a comprehensive framework with three evaluation steps and four evaluation tasks is designed.
    
    
    \item We construct a more comprehensive Chinese humor dataset compared with the prior research, which fully meets all the data requirements of the proposed evaluation framework.
    
    
    \item Our empirical study justifies the positive effect of fine-tuning and external knowledge for PLMs on humor-relevant tasks, which has important guiding value for future optimization of PLMs in humor understanding and generation.

\end{itemize}

\vspace{-0.5em}
\section{EVALUATION FRAMEWORK}
\begin{figure}[t]
    \centering
    \includegraphics[width=0.85\linewidth]{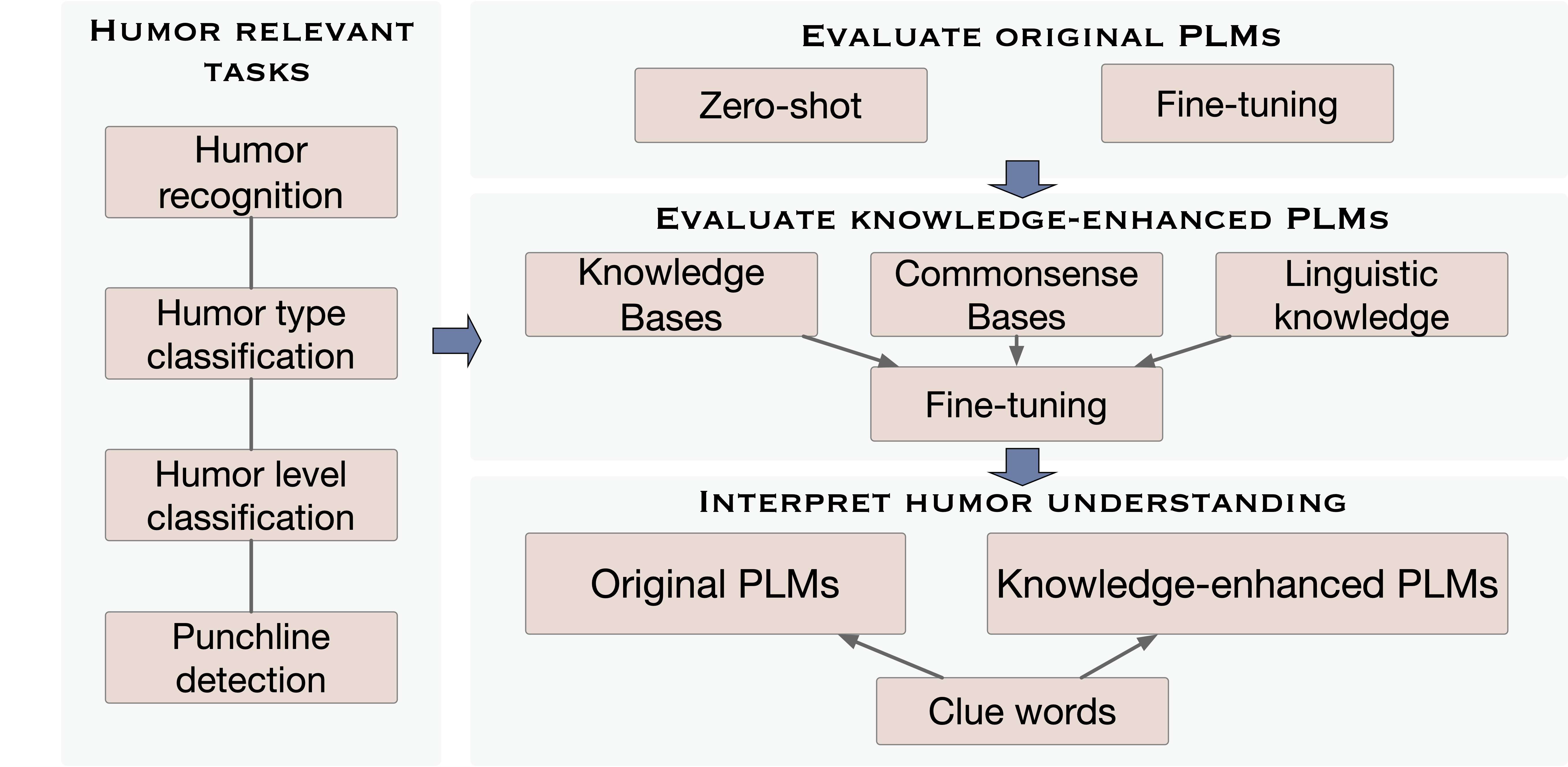}
    \caption{
    \small The evaluation framework of PLMs’ humor understanding, including four tasks and three steps.
    }
    \label{fig:pipeline}
    \vspace{-4mm}
\end{figure}

\vspace{-0.5em}
In this paper, we focus on evaluating the ability of PLMs in understanding humor. Only when PLMs are capable of understanding humor can they generate more reasonable humorous texts, so we leave the investigation of the humor generation ability of PLMs for future work.
As depicted in Fig.~\ref{fig:pipeline}, we propose a comprehensive evaluation framework with three evaluation steps based on four evaluation tasks to investigate the capability of PLMs in understanding humor. In the following, we introduce them in detail.

\vspace{-0.5em}
\subsection{EVALUATION TASKS}
\begin{figure*}[t]
    \centering
    \includegraphics[width=0.8\linewidth]{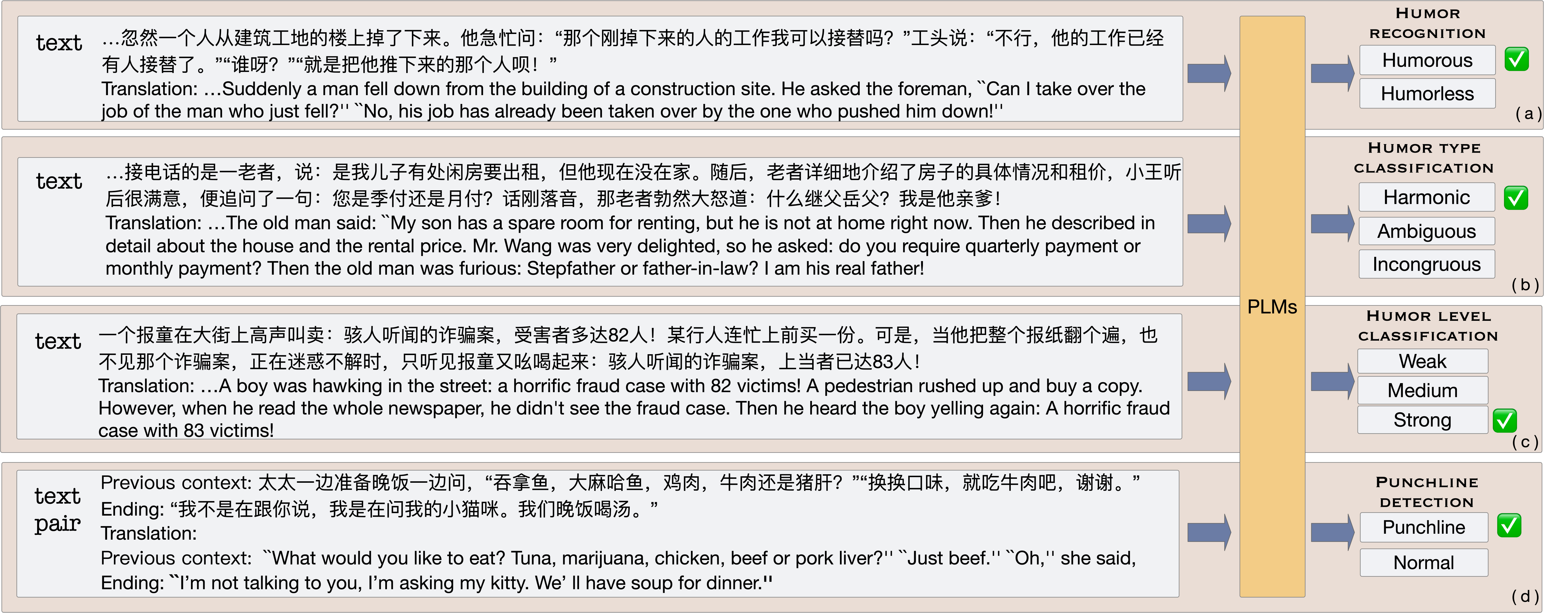}
    \caption{
    \small The process for PLMs to perform on four humor-relevant tasks.}
    \label{fig:in-out}
    \vspace{-4mm}
\end{figure*}

We evaluate four representative humor-relevant tasks as follows:

\noindent{\bf Humor Recognition}.
This task aims to distinguish humorous texts from humorless ones~\citet{Computationally_recognizing_wordplay}. For each input text, the task outputs \textit{humorous} or \textit{humorless} as shown in Fig~\ref{fig:in-out}(a). 

\noindent{\bf Humor Type Classification}. This task first appears in the CCL2019 competitions~\footnote{http://www.cips-cl.org/static/CCL2019/call-evaluation.html\label{bisai}}. Given a piece of humor text as input, it classifies humorous texts into several predefined humorous types and outputs \textit{harmonic}, \textit{ambiguous} or \textit{incongruous} as shown in Fig~\ref{fig:in-out}(b).

\noindent{\bf Humor Level Classification}. 
This task works on judging the level of humor for input texts, also first proposed in the CCL2019 competitions, where the humor can be divided into five levels from the weakest to the strongest. Here we modify the five continual levels into three discrete levels. Given a piece of humor text, the task outputs its corresponding humor level \textit{(strong, medium or weak)} as shown in Fig~\ref{fig:in-out}(c).

\noindent{\bf Punchline Detection}.
According to humor theory~\citep{Whats_SO_Bloody,A_Study_of}, this task determines whether there is a semantic incongruity between the previous context and its punchline (or laugh-point) ending, which originates but is slightly different from the research by~\citet{Predicting_Audiences_Laughter}.
Specifically, the input of this task is a pair of texts: 1) the context of a humorous text before the punchline ending sentence and 2) its corresponding punchline ending sentence or a non-punchline normal ending sentence, and it will determine whether this ending sentence is a punchline one as shown in Fig~\ref{fig:in-out}(d).

The above four tasks are adopted for different evaluation purposes: Initially, humor recognition is the most basic task to examine a model's ability in discriminating humor.
Further, humor type classification evaluates a model's ability in distinguishing between different types of humorous text, while humor level classification indicates whether the model is sensitive to different levels of humor.
Last but not the least, punchline detection can reflect whether a model has a fine-grained understanding on a humorous text and thus judges whether the ending sentence is a punchline or normal one.

\subsection{EVALUATION STEPS}

To perform a thorough and insightful evaluation based on the above four humor-relevant tasks, three evaluation steps are designed for investigating the ability of PLMs in humor understanding:

\noindent{\bf Evaluate Original PLMs.} The first step is responsible for evaluating the humor understanding ability of the original PLMs, which is expected to tell us: 1) Whether the original PLMs have humor understanding ability; and 2) What are the improvements and shortcomings of PLMs after simple fine-tuning on the humor dataset?

\noindent{\bf Evaluate Knowledge-enhanced PLMs.} The second evaluation step tries to inject different kinds of external knowledge into PLMs, which is to let us know: What kind of knowledge can help improve PLMs in humor understanding and to what extent? 


\noindent{\bf Interpret Humor Understanding.} The third step explores whether PLMs’ humor understanding ability is in line with human perception. For this purpose, it investigates whether the PLMs (including the original PLMs, fine-tuned PLMs, and knowledge-enhanced PLMs) can detect appropriate clue words from the humorous text in the concerned humor tasks.

\section{THE CHINESE HUMOR DATASET}

To fully meet all data requirements of the evaluation framework, we construct a large-scale Chinese humor dataset, which consists of four sub-datasets as depicted in Table~\ref{tab:data}.


\begin{table}[t]
\small
\caption{\small The overview of the constructed Chinese humor dataset used for four humor-relevant tasks.}
    \begin{center}
        \begin{threeparttable}
            \setlength{\tabcolsep}{1mm}{
            \resizebox{0.38\textwidth}{!}{
            \begin{tabular}{p{1.5cm}<{\centering}p{1.3cm}<{\centering}|p{1.8cm}<{\centering}p{1.1cm}<{\centering}}
                    \toprule
                    \multicolumn{2}{c|}{\bf Humor Recognition} &\multicolumn{2}{c}{\bf Punchline Detection}\\
                    \midrule
                    \bf Property & \bf Amount  & \bf Property & \bf Amount  \\
                    \midrule
                    Humorous &18709& Punchline &18709 \\
                    Humorless &7709&Normal& 18709 \\
                    \midrule
                    \midrule
                    \multicolumn{2}{c|}{\bf Humor Type Classification}& \multicolumn{2}{c}{\bf Humor Level Classification}\\
                    \midrule
                    \bf Property & \bf Amount  & \bf Property & \bf Amount  \\
                    \midrule
                    Harmonic&557&Weak&3722\\
                    Ambiguous&972&Medium&5764\\
                    Incongruous&2977&Strong&3009\\
                    \bottomrule
            \end{tabular}}}
        \end{threeparttable}
    \end{center}
    \label{tab:data}
    \vspace{-4mm}
\end{table}

\textbf{Humor Recognition Sub-dataset. }
The humor recognition sub-dataset contains humorous texts mainly from released data\textsuperscript{\ref{bisai},}\footnote{https://github.com/liuhuanyong/ChineseHumorSentiment\label{kaiyuan}} and humorless text
crawled from various platforms.
On another hand, to construct negative examples, we mainly crawl short monologues or dialogues, such as hint fictions, fables, celebrity stories, or monologue-based diaries as the humorless texts. Such texts have similar language styles with the humorous ones. The lengths of the texts are ranged from 20 to 200 characters. Each humorless text is tagged by three recruited human volunteers to evaluate whether this piece of text is actually humorless. Texts with controversial labels given by the three volunteers will be discarded.

\textbf{Humor Type Classification Sub-dataset. }
The humor type classification sub-dataset contains three types of humorous texts: 
\begin{itemize}
\item {\em Harmonic humor}: it means a word-pair has a similar pronunciation but different meanings in a piece of humorous text, such as 
\begin{CJK*}{UTF8}{gkai}
	``季付\ (meaning: pay quarterly, Chinese pinyin: jifu)''\ and\ ``继父\ (meaning: stepfather, Chinese pinyin: jifu)'',\  ``月付\ (meaning: pay monthly, Chinese pinyin: yuefu)''\ and\ ``岳父\ (meaning: father-in-law, Chinese pinyin: yuefu)''
\end{CJK*}
in the first row of Table~\ref{tab:htdd}
; 

\item {\em Ambiguous humor}: it means at least two definitions of a word are simultaneously used in a piece of humorous text. For example, 
\begin{CJK*}{UTF8}{gkai}
	``十分''
\end{CJK*}
can mean ``very'' and ``ten scores'' at the same time.
in the second row of Table~\ref{tab:htdd}
;

\item {\em Incongruous humor}: it means there is a semantic incongruity in a humorous text, which doesn’t follow humans’ expectation. For example, in the third row of Table~\ref{tab:htdd},
\begin{CJK*}{UTF8}{gkai}
	``你看这个姑娘就很有素质一直很冷静嘛''
\end{CJK*}
which means ``the girl is calm without dispute'', and the response is 
\begin{CJK*}{UTF8}{gkai}
	``我住30楼，跑下来累了歇会儿再骂你''
\end{CJK*}
which means ``the girl keeps calm because she is too tired to quarrel''. Such response has semantic incongruity which doesn’t follow humans' normal expectation.
\end{itemize}
%
We first collect the data from the release data\textsuperscript{\ref{bisai},\ref{kaiyuan}} which already have labels. In order to guarantee the accuracy of the type labels, we proofread and then discard the wrong ones with the help of the recruited three human raters.

\begin{table}[t]
\small
\caption{\small Human evaluation on normal endings generated by CPM based on semantics, correctness and readability.}
    \begin{center}
        \begin{threeparttable}
        \resizebox{0.38\textwidth}{!}{
            \begin{tabular}{p{0.9cm}|p{6cm}}
                \toprule
                \bf Criteria &(\bf Scores) \bf Contents\\
                \midrule
                \makecell[l]{Seman-\\tics} & \tabincell{l}{\makecell[l]{(5) Link very closely to the previous context.} \\
                \makecell[l]{(4) Link highly closely to the previous context.} \\
                \makecell[l]{(3) A majority part links to the previous context.} \\
                \makecell[l]{(2) A minority part of links to the previous context.} \\
                \makecell[l]{(1) Can’t link to the previous context completely.}}\\ 
                \midrule
                \makecell[l]{Correct-\\ness}  & \tabincell{l}{(5) Completely factually correct.\\
                (4) Highly factually correct.\\
                (3) A majority part is factually correct.\\
                (2) A minority part is factually correct.\\
                (1) Completely factually wrong.}\\ 
                \midrule
                \makecell[l]{Read-\\ability}  & \tabincell{l}{(5) Extremely readable without grammar mistakes.\\
                (4) Highly readable 1 grammar mistakes.\\
                (3) A majority part is fluent with a few grammar mistakes.\\
                (2) A minority part is fluent with many grammar mistakes.\\
                (1) Not readable at all.}\\ 
                \bottomrule
            \end{tabular}}
        \end{threeparttable}
    \end{center}
    \label{tab:scoresheet}
    \vspace{-4mm}
\end{table}

\begin{table*}[t]
\small
\caption{\small An example of the humor type classification sub-dataset, including harmonic humor, ambiguous humor and incongruous humor. The underlined part is the clues indicating a specific type.}
    \begin{center}
        \begin{threeparttable}
            \resizebox{0.89\textwidth}{!}{
            \begin{tabular}{p{1.5cm}|p{5.9cm}|p{7.6cm}|p{0.9cm}<{\centering}}
                    \toprule
                    \bf Property & \bf\centering Content (Chinese)  &\bf\centering Content (English)  & \bf Length \\
                    \midrule
                    Harmonic &  
                    \begin{CJK*}{UTF8}{gkai}
                    	...您是\underline{季付}还是\underline{月付}？什么\underline{继父}\underline{岳父}？我是他亲爹！
                    \end{CJK*}
                    &
                    ...do you require \underline {quarterly payment} or \underline {monthly payment}? \underline {Stepfather} or \underline {father-in-law}? I am his real father!
                    &162 \\
                    \midrule
                    Ambiguous &
                    \begin{CJK*}{UTF8}{gkai} ...\underline{十分}简单。...\underline{十分}简单，剩下九十分很难！
                    \end{CJK*}
                    &
                    ...\underline {Very} easy. ...\underline {ten points} are easy, the ninety points are difficult!
                    &116\\
                    \midrule
                    Incongruous &
                    \begin{CJK*}{UTF8}{gkai} ...“\underline{你看这个姑娘就很有素质一直很冷静}
                    \underline{嘛}。”“\underline{我住30楼，}
                    \underline{跑下来累了歇会儿再骂你}。”
                    \end{CJK*}
                    &
                    ...``\underline{Look at this girl, she is very calm.'' ``I need a rest after running} \underline{down from the 30$^{th}$ floor and then to scold you}.''
                    &125\\
                    \bottomrule
            \end{tabular}}
        \end{threeparttable}
    \end{center}
    \label{tab:htdd}
    \vspace{-4mm}
\end{table*}

\textbf{Humor Level Classification Sub-dataset.}
The humor level classification sub-dataset classifies humorous texts into three levels of humor: weak, medium, and strong. 
We collect the data from the same places as the humor type classification sub-dataset, and also perform similar manual proofreading operations to guarantee the accuracy of the labels.
%

\textbf{Punchline Detection Sub-dataset. }
Based on the humorous text in humor recognition sub-dataset, we construct humorous and humorless context-ending pairs.

The humorous ending is extracted by dividing each humorous text into two parts: previous context and punchline. 
According to the theories of humor~\citep{The_ancient_roots,Humor,Cognitive_aspects_of,Whats_SO_Bloody,A_Study_of}, the reason why humor introduces laughter is that a piece of text presents an unexpected sentence which is incongruous with the previous context. Thus, we extract the unexpected sentence in a humorous text as the punchline with the help of human annotation.
Specifically, we first enroll another three volunteers, and each of them is required to vote punchline sentences for all humorous texts. We then choose the sentence which has the highest vote as the final punchline sentence for each humorous text and discard the following content in this piece of text. If more than one sentences have equal high votes, we discard this piece of humorous text. 

The normal ending is generated by a text generator. We input the previous context into a large language model such as CPM~\citep{CPM,CPM-2}, and generate a normal ending which has similar length with the punchline.
We also carry out human evaluation and machine evaluation to guarantee the quality of the generated normal endings.
The steps of quality evaluation are as follows:
i) We first enroll another three volunteers and randomly select 3,000 normal endings. Each of the volunteers needs to give a rating for the overall 3,000 normal endings based on a scoresheet shown in Table~\ref{tab:scoresheet}. 
We calculate Inter-rater agreement of Krippendorff’s Alpha (IRA) to ensure the confidence of human ratings. For the controversial ratings which have low agreements ($<$0.7) or a normal ending is rated below 0.85, we re-generate a new normal ending. 
ii) Next, we use simCSE~\citep{SimCSE} to calculate similarity scores to guarantee the generated normal endings have similar semantics with the corresponding punchlines and the previous contexts.
The similarity score between the normal ending and the punchline ending is $s_1$, and the similarity score between context-punchline pair and context-normal-ending pair is $s_2$. The final similarity score $s$ is the average of $s_1$ and $s_2$.
For some normal endings whose $s$ are below 0.85, we also re-generate new ones.

\vspace{-0.5em}
\section{METHODOLOGY}
\begin{figure*}[t]
    \centering
    \includegraphics[width=0.82\linewidth]{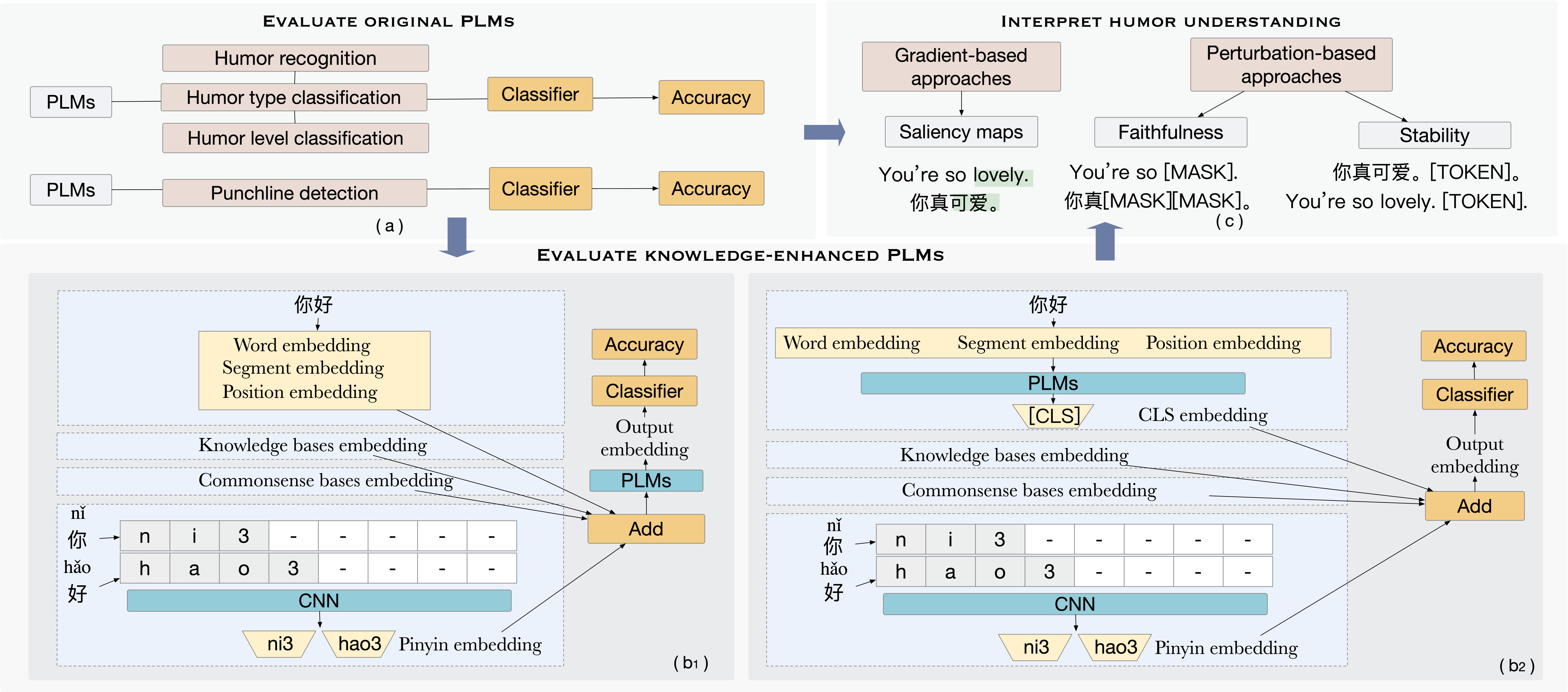}
    \caption{
    \small The evaluation framework of PLMs’ humor understanding, including three steps: evaluate original PLMs (a), evaluate knowledge-enhanced PLMs (b$_1$,b$_2$), and interpret humor understanding (c).
    }
    \label{fig:framework}
    \vspace{-4mm}
\end{figure*}

\vspace{-0.5em}
Fig~\ref{fig:framework} illustrates our framework for evaluating PLMs’ ability in understanding humor, which consists of three parts: evaluate original/fine-tuned PLMs, evaluate knowledge-enhanced PLMs, and interpret humor understanding in PLMs.

\subsection{EVALUATE ORIGINAL/FINE-TUNED PLMS}
In this module, we adopt several SOTA PLMs to model humor understanding through four representative tasks: humor recognition, humor type classification, humor level classification, and punchline detection (see Fig~\ref{fig:framework}(a)).
The first three are text classification tasks that take a piece of text as input and output text properties (i.e., humorous or humorless, humor types, and humor levels).
The last task is performing text matching of a context-ending pair, and outputs the similarity of the pair to indicate whether the ending is a punchline or not.
We calculate the similarity scores based on sentence-level embeddings, and the loss function we adopt is online contrastive loss~\footnote{https://www.sbert.net/}, which is proved better than contrastive loss~\citep{Dimensionality_Reduction_by} in our experiments.
If the similarity of the pair is over 0.5, we regard the ending as a normal ending. Otherwise, we regard the ending as a punchline.

\subsection{EVALUATE KNOWLEDGE-ENHANCED PLMS}
In this module, we consider several types of external knowledge, such as general knowledge bases, commonsense bases, and linguistic knowledge. 
For each type of external knowledge as shown in Fig~\ref{fig:framework}(b), we further use two ways of knowledge enhancement, i.e., \textit{implicit embedding} and \textit{explicit fusion}, to enhance PLMs. 

\textbf{Knowledge Embedding Construction. }
We utilize open-sourced Tencent AI Lab Embedding Corpus~\footnote{https://ai.tencent.com/ailab/nlp/zh/embedding.html} and ConceptNet~\footnote{https://github.com/commonsense/conceptnet-numberbatch} as knowledge embeddings from general knowledge bases and commonsense bases, respectively.
For the linguistic knowledge, we learn a pinyin embedding for each character as the knowledge embedding to detect different semantic meanings for characters with the same or similar pronunciations.
We first utilize the \emph{pypinyin}~\footnote{https://pypi.org/project/pypinyin/} package to generate pinyin with one of four tones for each Chinese character in a given text. For polyphonic characters, we select the first pronunciation. 
Inspired by~\citet{ChineseBERT}, we use special tokens to denote tones.
The maximum length of input pinyin sequence is set as 8 and we use a special letter ``-'' for padding short pinyin sequences. 
We adopt a Convolution Neural Network model~\citep{Backpropagation_applied_to} to make pinyin embedding as shown below: 
\begin{small}
\begin{gather}
\vspace{-2em}
\label{eq:cnnpinyin}
    Emb_{pinyin}=Maxpool(CNN(pypinyin(Seq_{in})))
\vspace{-2em}
\end{gather}
\end{small}








\textbf{Fusion Layer. }
After constructing three types of knowledge embeddings, we normalize each of the embeddings by aggregating along the token dimension of each word or letter dimension of each pinyin (as each word is a sequence of tokens and each pinyin is a sequence of letters), respectively, with a fully connected layer to get normalized knowledge embedding(s) $\overline{Emb}_{k}$. 
Then we use two ways of knowledge enhancement for PLMs, which are Implicit embedding and Explicit fusion.
Specifically, if we use BERT as the PLM, implicit embedding is to add word embedding $Emb_{word}$, segment embedding $Emb_{seg}$, position embedding $Emb_{pos}$ and normalized knowledge embedding $\overline{Emb}_{k}$ together as the input embeddings for PLMs. The output embeddings for PLMs are used to make new predictions. The process is shown as follows:
\begin{small}
\begin{gather}
\vspace{-2em}
    Emb_{in}=Emb_{word}+Emb_{seg}+Emb_{pos}+\overline{Emb}_{k}\\
    Emb_{out}=PLM(Emb_{in})\\
    y=\bm{W}*Emb_{out}+b 
\vspace{-2em}
\end{gather}
\end{small}

Besides, explicit fusion is to enhance the sentence embedding given by a PLM with a fusion layer. 
It's well-known that the embedding of the \texttt{[CLS]} token in BERT-style PLMs represents the information of a whole sentence.
We add the normalized knowledge embedding $\overline{Emb}_{k}$ to the PLMs’ sentence embedding $Emb_{sen}$ before the head of the output to make new predictions. The process is as follows:
\begin{small}
\begin{gather}
\vspace{-2em}
    Emb_{in}=Emb_{word}+Emb_{seg}+Emb_{pos}\\
    Emb_{sen}=PLM(Emb_{in})\\
    Emb_{out}=Emb_{sen}+\overline{Emb}_{k}\\
    y=\bm{W}*Emb_{out}+b 
\vspace{-2em}
\end{gather}
\end{small}
\noindent where \bm{$W$} and $b$ are trainable parameter and bias, respectively.


\subsection{INTERPRET HUMOR UNDERSTANDING}
In this module, given an input text, we aim at figuring out which input words are critical for a PLM to make it correctly perform humor-relevant tasks and whether these words better interpret PLMs' humor understanding ability~\citep{lyu2022multimodal,lyu2023attention}.
We utilize gradient-based and perturbation-based techniques with the Captum package~\footnote{https://captum.ai/} for this purpose (see Fig.~\ref{fig:framework}(c)).

Gradient-based approaches~\citep{Deep_Inside_Convolutional} compute saliency map based on the gradient of the input with respect to the output. 
We first take 
differentiable embeddings of tokens as the input for PLMs.
Next, we aggregate the embeddings gradients with L2 normalization. Then we use Input X Gradient~\citep{Learning_Important_Features}, which multiplies the gradient with the normalized embeddings, to improve the sharpness of the saliency scores, thus to compare them better.
After that, we use 
a visualizer to present saliency maps for saliency scores
to find clue words for PLMs' correct predictions on humor-relevant tasks.

Perturbation-based techniques perturb the input to find which input regions have a significant impact on the prediction. 
Given an input text, we randomly perturb it by adding or removing a few tokens and further check the new prediction of the PLMs.
Based on perturbation-based techniques, we first investigate the faithfulness of the saliency maps to detect whether PLMs’ correct prediction 
are not based on arbitrary choices.
We replace the top-$N$ (we set $N=3$) most salient words with a mask token and then measure the drop of the PLMs’ performance.
Next, we investigate the stability of the saliency maps to detect whether insignificant words affect saliency maps.
We add some random words at the end of the texts and then measure the correlation between the change of the prediction and the change of the saliency scores based on the Pearson correlation coefficient
and Spearman correlation coefficient.

\vspace{-0.5em}
\section{EXPERIMENTS}
\vspace{-0.5em}
Following the proposed evaluation framework, we carry out experiments to investigate whether the pre-trained language models (PLMs) have the ability of humor understanding (see Sec.~\ref{results-original}), whether external knowledge can improve their humor understanding ability (see Sec.~\ref{results-knowledge}), and the interpretability of the detected clue words which lead to PLMs’ correct prediction on humor understanding tasks (see Sec.~\ref{results-interpretability}). 
%

\textbf{Experiment Setup. }Our experiments are carried on GeForce RTX 3090 GPU (on our machine) and TPU (on Google Colab) with Pytorch in Python.
The sequence length is set to 200.
We initialize the learning rate to 2e-5 and batch size from 4 to 32 according to the memory of the machine, and use early stopping with 20 epochs.

\textbf{Baselines, Datasets and Metrics. }
We adopt some representative PLMs, including base and large versions of BERT~\citep{BERT}, RoBERTa~\citep{RoBERTa}, BART~\citep{BART}, T5~\citep{T5}, CPT~\citep{CPT} for humor recognition, humor type classification, and humor level classification tasks.
For the punchline detection task, we adopt base and large versions of simCSE-BERT~\citep{SimCSE} and simCSE-RoBERTa~\citep{SimCSE}, which are proved to perform better in text matching.
We divide each dataset into a training set and a dev set at a ratio of 7 to 3, and make down-sampling for all training and dev set to balance sample numbers in different classes. 
We use accuracy with percentage as the metric.

\subsection{RESULTS ON ORIGINAL/FINE-TUNED PLMS}
\label{results-original}
The evaluation results on original and fine-tuned PLMs based on four humor-relevant tasks are shown in Table~\ref{tab:re-hr} (see the column ``zs'' and ``ft'').
From the results, we observe that the original PLMs have weak ability on humor understanding with the average value 54.98, 33.77, 33.32, and 49.96 in the zero-shot learning on the humor recognition, humor type classification, humor level classification, and punchline detection task, respectively.
After fine-tuning on the corresponding sub-datasets, the performance is improved by 68.67\%, 72.64\%, 40.87\%, and 91.89\%, respectively. The accuracy on humor recognition and punchline detection are both over 90\%. It suggests that \emph{PLMs have a certain degree of ability in humor recognition and punchline detection after fine-tuning on the humor dataset}.


\begin{table*}[t]
\small
\caption{\small The evaluation results on original PLMs and knowledge-enhanced PLMs based on humor recognition (Column 2 to 4), humor type classification (Column 5 to 7), humor level classification (Column 8 to 10), and punchline detection (Column 11 to 13). zs: zero-shot learning, ft: fine-tuning, K-ft: knowledge-enhanced fine-tuning. The improvement rate is to compare with column ``zs''.}
    \begin{center}
        \begin{threeparttable}
        \resizebox{0.89\textwidth}{!}{
            \begin{tabular}{p{1.9cm}|p{0.5cm}p{0.8cm}p{0.8cm}|p{0.5cm}p{0.8cm}p{0.8cm}|p{0.5cm}p{0.8cm}p{0.8cm}||p{2.1cm}|p{0.5cm}p{0.8cm}p{0.8cm}}
                \toprule
                \multirow{2}{2cm}{\bf PLMs} & \multicolumn{3}{c|}{\bf Humor recognition} &
                \multicolumn{3}{c|}{\bf Humor type classification} &
                \multicolumn{3}{c||}{\bf Humor level classification}&\multirow{2}{2.2cm}{\bf PLMs} & \multicolumn{3}{c}{\bf Punchline detection}  \\
						&\bf zs&\bf ft&\bf K-ft
						&\bf zs&\bf ft&\bf K-ft
						&\bf zs&\bf ft&\bf K-ft
						&
						&\bf zs&\bf ft&\bf K-ft\\
                \midrule
                BERT-base & 51.08 &92.21&\bf93.17&33.50&60.14&\bf62.36&34.98& 46.35& \bf47.24&S-BERT-base & 49.27&95.21&\bf96.30\\
                BERT-large  &51.69 &92.87&\bf94.32&34.35&61.04&\bf64.31&35.03& 47.36& \bf48.49&S-BERT-large  &50.01 &95.24&\bf96.7\\
               RoBERTa-base & 52.83 &91.79&\bf93.46&32.51&61.42&\bf63.50&36.13& 45.45& \bf46.37&S-RoBERTa-base & 52.43 &96.51&\bf97.88\\
            RoBERTa-large & 52.96 &92.30&\bf93.24&33.16&62.58&\bf64.01&37.11& 46.26& \bf47.02&S-RoBERTa-large & 48.13 &96.52&\bf97.98\\
                BART-base & 49.35 &91.46&\bf92.58&29.40&46.23&\bf47.01&31.09& 46.22&\bf46.80&-&-&-&-\\
               BART-large & 50.06 &93.21&\bf93.89&33.73&48.32&\bf49.10&32.07&47.25&\bf47.97&-&-&-&-\\
                T5-base& 52.05 &91.47&\bf92.45&33.32&55.10&\bf56.22&31.28& 46.82&\bf47.96&-&-&-&-\\
                T5-large& 54.26 &93.19&\bf93.99&35.28&57.87&\bf58.86&32.09& 47.51&\bf48.76&-&-&-&-\\
                CPT-base & 66.21 &94.35&\bf95.85&34.37&64.91&\bf66.30&31.24& 47.88&\bf49.24&-&-&-&-\\
                CPT-large & 69.27 &94.45&\bf95.91&38.04&65.33&\bf67.60&32.20& 48.30&\bf50.21&-&-&-&-\\
                \midrule
                Average&54.98&	92.73&\bf	93.89&	33.77&	58.29&\bf	59.93&	33.32&	46.94&\bf	48.01&Average&49.96&	95.87&	\bf97.24\\
                Improve rate&-  &68.67\%&\bf70.78\%&-&72.64\%&\bf77.47\%&-&40.87\%&\bf44.07\%&Improve rate&-&91.89\%&\bf94.63\%\\
                \bottomrule
            \end{tabular}}
        \end{threeparttable}
    \end{center}
    \label{tab:re-hr}
    \vspace{-4mm}
\end{table*}


\subsection{RESULTS ON KNOWLEDGE-ENHANCED PLMS}
\label{results-knowledge}
In this part, we only present the experimental results by injecting Chinese pinyin into PLMs in the way of explicit fusion as shown in Table~\ref{tab:re-hr}. 
That is because we observe in our experiments that either injecting Chinese pinyin in the way of implicit embedding, or injecting another one or two types of knowledge embeddings in any ways do not improve PLMs' performance in all the four humor-relevant tasks. 
Due to space limitation, we omit these results and will make analysis later.

See the column ``K-ft'' in the above Tables, we find that PLMs perform better, improving by 70.78\%, 77.47\%, 44.07\%, and 94.63\% on the humor recognition, humor type classification, humor level classification, and punchline detection task respectively.
%
%
This group of experiments demonstrates that \emph{external linguistic knowledge such as Chinese pinyin has a positive effect for PLMs in humor-relevant tasks}, and injecting external knowledge by explicit fusion is more possible to maintain important information in the knowledge than by implicit embedding.



However, for another one or two types of knowledge, which do not improve the performance of PLMs in all the four evaluation tasks, we give the possible reasons as follows:
1) The huge amount of data used for training PLMs may already contain most of the factual knowledge and commonsense knowledge, thus the existing knowledge bases or commonsense bases can not contribute more for PLMs in humor understanding.
2) Some humorous texts need complicated specific knowledge or inference paths to understand, which can not be provided by existing knowledge. For example, in the following harmonic humorous text
\begin{CJK*}{UTF8}{gkai} 
“...一来闻，二来闻，三来闻..”, the Chinese phrase “一来闻”
\end{CJK*}
has a similar pronunciation with the English word ``eleven'',  where ``e'' and ``leven'' correspond the pronunciation of Chinese character 
\begin{CJK*}{UTF8}{gkai} 
``一'' and ``来闻'', respectively.
\end{CJK*}
Therefore, 
\begin{CJK*}{UTF8}{gkai} 
``二来闻 (two-leven)'' and ``三来闻 (three-leven)'', 
\end{CJK*}
are analogous to 
\begin{CJK*}{UTF8}{gkai} 
``一来闻 (one-leven)'',
\end{CJK*}
, which produce harmonic humor. It’s a much difficult inference process for PLMs to understand and make correct humor type classification.

Moreover, when we inject other existing knowledge from knowledge bases and commonsense bases except Pinyin knowledge in the way of implicit embedding into PLMs, the performance for PLMs in the humor-relevant tasks do not have any improvement. Due to space limitation, we also omit these results in our paper and give some possible analysis as follows:
1) Fusing Knowledge from different sources will do harm to separate feature of each type of knowledge. It's difficult for PLMs to learn effective information from the fused knowledge.
2) Humor is a much tough issue. The existing knowledge is not powerful enough for PLMs in humor understanding. 
Thus, \emph{besides linguistic knowledge, PLMs also need humor-relevant background knowledge for better performance in humor-relevant tasks and better humor understanding ability}.

\vspace{-0.5em}
\subsection{RESULTS ON INTERPRETABILITY ANALYSIS}
\label{results-interpretability}
To visualize the interpretability of PLMs' humor understanding ability, we draw saliency maps for sentences to show the detected clue words.
We first investigate the stability of the clue words detection results to verify whether these clue words are faithful to PLMs. We add some random characters such as 
\begin{CJK*}{UTF8}{gkai} 
“我 (I)”
\end{CJK*}
at the ending position of samples, and then measure the correlation between the change in the prediction and the change in the saliency scores. The p-values of Pearson correlation coefficient and Spearman correlation coefficient are 0.0087 and 0.0010, respectively in the zero-shot learning, 0.0086 and 0.0060, respectively in the fine-tuning, and 0.0080 and 0.0064, respectively in the knowledge-enhanced fine-tuning. The changes are statistically different due to p-values all below 0.05, which suggests that saliency scores highly relate to predictions, thus the saliency scores are stable and the clue words detection results are faithful to PLMs, which can be trusted by humans for interpreting PLMs' humor understanding ability.

\begin{table*}[t]
\small
\caption{\small The results of masking the top three salient words except \texttt{[SEP]} of each instance based on the zero-shot learning, fine-tuning in humor recognition (Column 2 to 3), humor type classification (Column 4 to 5), humor level classification (Column 6 to 7), and punchline detection  (Column 9 to 10). mzs: mask words in the zero-shot learning, mft: mask words in the fine-tuning, mkft: mask words in the knowledge-enhanced fine-tuning. The decline is to compare with the corresponding columns in Table~\ref{tab:re-hr}.
}
    \begin{center}
        \begin{threeparttable}
        \resizebox{0.89\textwidth}{!}{
            \begin{tabular}{p{2.2cm}|cc|cc|cc||p{2.2cm}|cc}
                \toprule
                \multirow{2}{2.2cm}{\bf PLMs} & \multicolumn{2}{c|}{\bf humor recognition} &
                \multicolumn{2}{c|}{\bf Humor type classification} &
                \multicolumn{2}{c||}{\bf Humor level classification}&\multirow{2}{2.2cm}{\bf PLMs}&\multicolumn{2}{c}{\bf Punchline detection} \\
						&\bf mzs&\bf mft
						&\bf mzs&\bf mft
						&\bf mzs&\bf mft
						&
						&\bf mzs&\bf mft\\
                \midrule
                BERT-base & 50.67& 52.35&32.19& 30.02& 33.21&  32.48&S-BERT-base &48.21&47.35\\
                BERT-large  &49.42&49.38&33.11&34.37&33.22&34.57&S-BERT-large  &48.29&50.52\\
               RoBERTa-base  &51.92&50.66&30.98&30.99&34.75& 34.21&S-RoBERTa-base  &51.33&52.01\\
            RoBERTa-large &50.27&51.33&32.67&33.01&34.55& 33.17&S-RoBERTa-large  &47.09&46.50\\
                BART-base  &48.65&47.36&27.58&28.02&30.23& 30.53&-  &-&-\\
               BART-large  &49.25&46.98&32.37&33.21&30.59&32.04&-  &-&-\\
                T5-base &50.22&51.34&32.38&31.72&30.69&31.11&-  &-&-\\
                T5-large &53.21&52.89&34.33&33.09&31.19&30.08&-  &-&-\\
                CPT-base  &64.30&64.17&32.88&32.74&30.58&29.29&-  &-&-\\
                CPT-large  &68.17&69.32&36.59&37.21&31.07&32.51&-  &-&-\\
                \midrule
                Average &53.61&53.58&32.51&32.44&32.01&32.00&Average &48.73&49.10\\
                Decline rate&2.55\%&73.07\%&3.87\%&79.71\%&4.11\%&46.69\%&Decline rate&2.52\%&96.27\%\\
                \bottomrule
            \end{tabular}}
        \end{threeparttable}
    \end{center}
    \label{tab:faith_hr}
    \vspace{-4mm}
\end{table*}

\begin{figure*}[t]
    \centering
    \includegraphics[width=0.65\linewidth]{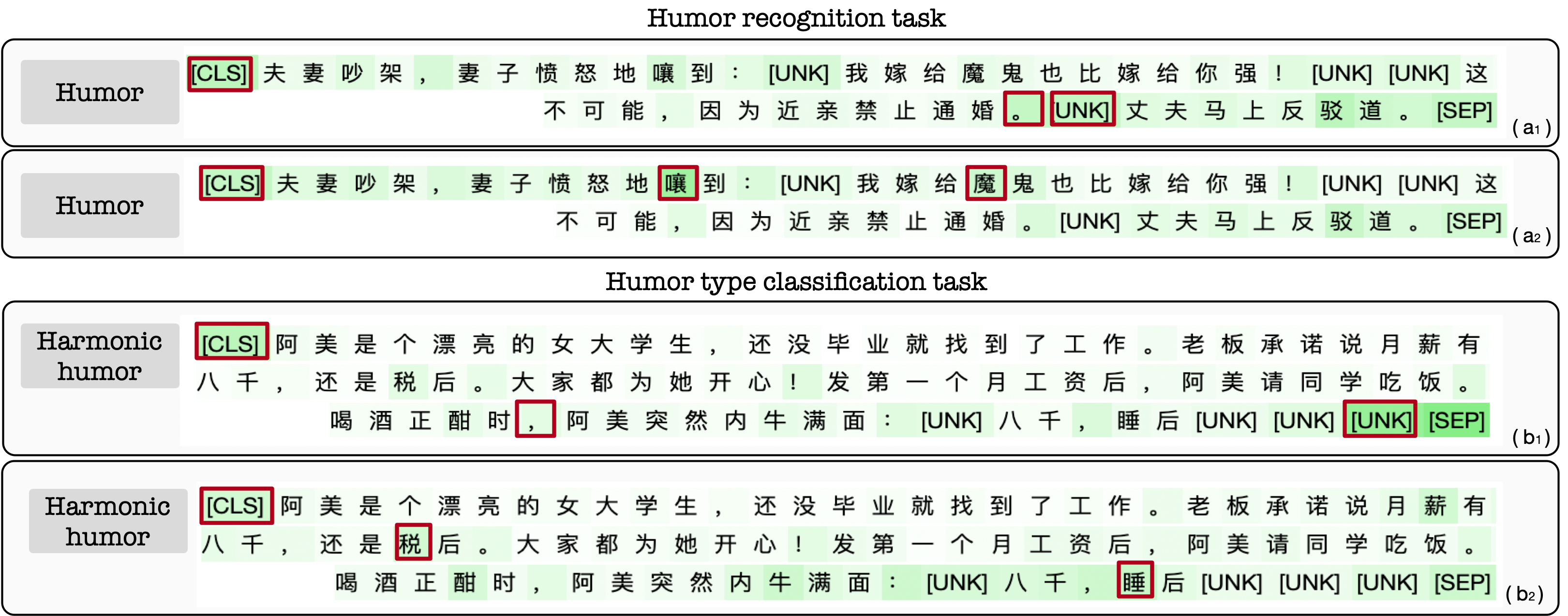}
    \caption{
    \small Saliency maps of some samples on the humor recognition sub-dataset based on the zero-shot learning (a$_1$) and fine-tuning (a$_2$), humor type classification sub-dataset based on the zero-shot learning (b$_1$) and fine-tuning (b$_2$). 
    The top three salient words except \texttt{[SEP]} are highlighted with red boxes.  We use BERT-base in these samples.
    }
    \label{fig:keshihua}
    \vspace{-4mm}
\end{figure*}

Fig~\ref{fig:keshihua} gives the saliency maps of several samples got from BERT-base humor-relevant tasks, and the faithfulness of the saliency map on four humor-relevant tasks is shown in Table~\ref{tab:faith_hr}.
In the humor recognition task, we observe that the original PLMs only take some special tokens, such as \texttt{[CLS]}, \texttt{[UNK]}, or punctuations (see Fig~\ref{fig:keshihua} (a$_1$)) as the clue words (which have deeper color).
%
We mask the top three salient words except \texttt{[SEP]} of each instance, and the average performance has a slight drop from 54.98 to 53.61.
In another three humor-relevant classification tasks, 
the results in the zero-shot learning as shown in
Fig~\ref{fig:keshihua} (b$_1$) are similar with those on the humor recognition task. 
These results prove that the original PLMs (without fine-tuning) can hardly understand humor, and their predictions on humor are nearly from arbitrary choices.

After that, we fine-tune PLMs on four humor-relevant tasks, respectively. We observe that on the humor recognition sub-dataset, the saliency maps for fine-tune PLMs show that the models focus more on the sentiment words, such as
\begin{CJK*}{UTF8}{gkai} 
“嚷 (shout)”
\end{CJK*}
,
\begin{CJK*}{UTF8}{gkai} 
“魔 (demon)”
\end{CJK*}
,
when making correct predictions (see Fig~\ref{fig:keshihua} (a$_2$)). 
We also mask the top three salient words
of each instance, and the average performance has a dramatic drop from 92.73 to 53.58.
When we fine-tune PLMs on humor type classification sub-dataset, PLMs focus more on the significant words, such as 
\begin{CJK*}{UTF8}{gkai} 
“税(meaning: taxes, pinyin: shui)”
\end{CJK*}
,
\begin{CJK*}{UTF8}{gkai} 
“睡(meaning: sleep, pinyin: shui)”
,
\end{CJK*}
in the harmonic humorous text as shown in Fig~\ref{fig:keshihua} (c$_2$),
\begin{CJK*}{UTF8}{gkai} 
“药(drug)”
\end{CJK*}
,
\begin{CJK*}{UTF8}{gkai} 
“吊(use)”
\end{CJK*}
,
in the ambiguous humorous text, and 
\begin{CJK*}{UTF8}{gkai} 
“考(have an examination)”
\end{CJK*}
,
\begin{CJK*}{UTF8}{gkai} 
“英(English)”
\end{CJK*}
,
in the incongruous humorous text when making correct predictions. We conjecture that \emph{PLMs find deep semantic correlations among these clue words, which help them make correct predictions.}
Moreover, when we fine-tune PLMs on the punchline detection sub-dataset, PLMs extract some significant words such as
\begin{CJK*}{UTF8}{gkai} 
“校(meaning: school, pinyin: xiao)”
\end{CJK*}
,
\begin{CJK*}{UTF8}{gkai} 
“孝(meaning: filial, pinyin: xiao)”
\end{CJK*}
,
in the previous context and ending, respectively, which lead to correct predictions for a punchline ending. 
However, these clue words are still not very apparent and some punctuations are also regarded as salient words
, which may interpret bad performance with serious threshold (0.8/0.2).
Therefore, \emph{fine-tuned PLMs have a certain degree of humor understanding ability after being fine-tuned on humor datasets. They focus on some significant words, such as sentiment words, which are partly in line with human perception on humor.}

\section{EVALUATION ON DOWNSTREAM TASKS}
\label{downstream}
Similar to happiness, sadness, and anger, humor is one of the essential emotions of humans. Therefore, we carry out further evaluations to investigate whether PLMs which have been fine-tuned on the humor dataset can achieve better performance on the downstream tasks.
We choose four Chinese sentiment classification datasets~\footnote{https://github.com/SophonPlus/ChineseNlpCorpus/raw/\\master/datasets/}: 
1) ChnSentiCorp-htl-all (Chn), which is a hotel review dataset with more than 5,000 positive reviews and more than 2,000 negative reviews; 2)
Waimai-10k (Wai), which contains 4,000 positive and 8,000 negative user reviews; 3) Online-shopping-10-cats (Shop), which has 30000 positive and 30000 negative user comments with online shopping; 4) weibo-senti-100k (Wei), which has about 50000 positive and 50000 negative comments.

We fine-tune BERT, which has been fine-tuned on our constructed Chinese humor dataset, including all the four sub-datasets. The results are shown in Table~\ref{tab:sendatasets}. The baseline results are based on BERT, which are derived from their published research~\citep{Scalable_multichannel_dilated, Takeaway_Comments_Sentiment, LongText_Sentiment_Analysis, IAS-BERT}.
We observe that the performance increases a little after fine-tuning on the humor dataset (see the row ``ft''). The results suggest that \emph{the sense of humor has common characteristics with other emotions, which thus humor-fine-tuned PLMs have a positive effect on other sentiment analysis tasks.}
We also find that there is no further improvement
after injecting external knowledge in any above-mentioned ways
(see the row ``K-ft''). It suggests that \emph{Chinese pinyin is a unique and important characteristic for humor understanding, 
which is not very useful for other relevant downstream tasks.
}  

\begin{table}[t]
\small
\caption{\small The accuracy of baselines, fine-tuned PLMs, and knowledge-enhanced fine-tuned PLMs on four Chinese sentiment classification datasets. The improvement is to compare with the baselines.
}
    \begin{center}
        \begin{threeparttable}
        \resizebox{0.45\textwidth}{!}{
            \begin{tabular}{l|cccc|cc}
                \toprule
                \bf Methods &\bf Chn&\bf Wai&\bf Shop&\bf Wei&\bf Average&\bf Improve rate\\
                \midrule
                baseline &93.32&92.42&93.20&97.90&94.21&- \\
                ft  &\bf94.21&\bf93.87&\bf93.66&\bf97.91&\bf94.91&\bf0.74\%\\
              K-ft  &94.20&92.98&93.76&97.88&94.70&0.52\%\\
                \bottomrule
            \end{tabular}}
        \end{threeparttable}
    \end{center}
    \label{tab:sendatasets}
    \vspace{-4mm}
\end{table}

\section{RELATED WORK}
PLMs demonstrate terrific capabilities on various downstream tasks~\citep{chen2024temporalmed,chen2023hadamard,chen2023hallucination,chen2022grow,chen2024talk,chen2023xmqas,chen2023mapo}.
We list two main related work as follows:

\textbf{Humor Datasets and Corpora. } 
Some research dedicate themselves to construct a large-scale humor datasets and corpora. 
For example, \citet{humor_norms_of} design a humor dataset which provides researchers with a list of humor ratings with 4,997 English words.
\citet{HAHA_2019_Dataset} present the development of a corpus of 30,000 Spanish tweets that were crowd-annotated with humor value and funniness score.
\citet{President_Vows_to} introduce a new dataset called Humicroedit
that design simple replacement edits to make English news headlines funny. 
\citet{UR-FUNNY} introduce a multimodal English humor dataset to detect 
humorous expressions in TED talks.
Different from the above research, we construct a Chinese humor dataset, which includes four sub-datasets, each of which can be used for one representative humor-relevant task.

\textbf{Humor Recognition. }Other research on humor focus on humor recognition which is to decide whether a given sentence expresses a certain degree of humor.
For example, \citet{Computational_Laughing} 
report text classification techniques are a viable approach to recognize humorous one-liners. 
\citet{Automatic_Detection_of} 
design several linguistic features
to automatically detect irony and humor in twitter.
\citet{Recognizing_Humour_using} adopt the minimum, maximum, and average Word2Vec similarity between ordered word pairs to recognize humor and extract humor anchor.
\citet{Humor_Recognition_and} investigate the latent semantic structures behind humor in four aspects.
\citet{Identifying_Humor_in} propose a generative language model 
and design some key component
to identify humor in reviews.
\citet{Predicting_Audiences_Laughter} 
use semantic structural features and semantic distance features to predict audience’s laughter in TED Talk Data based on convolutional Neural Network.
However, most of them design linguistic features to recognize humor. They ignore the powerful learning abilities of pre-trained language models (PLMs). Different from them, we design a comprehensive evaluation framework to make a research on PLMs’ humor understanding ability.

\section{CONCLUSIONS AND FUTURE WORK}
Humor understanding for PLMs is a challenging research in Natural Language Processing.
In this work, we systematically investigate the humor understanding ability of PLMs with a designed comprehensive framework with three evaluation steps and four evaluation tasks.
We also construct a comprehensive Chinese humor dataset, and our empirical study on it yields some valuable observations
:
1) While the original PLMs can hardly understand humor, they could gain a certain degree of humor understanding ability through fine-tuning.
2) Linguistic knowledge such as Chinese Pinyin has a positive effect for PLMs in humor-relevant tasks.
3) The existing knowledge bases and commonsense bases can not provide much required knowledge for humor understanding.
%
%
%
As a future work, we will find ways to collect more cultural knowledge for the optimization of PLMs in humor understanding.



%

\section{ACKNOWLEDGEMENT}
This work was supported by Science and Technology Commission of Shanghai Municipality Grant (No. 22511105902), National Natural Science Foundation of China (No.62072323), Shanghai Municipal Science and Technology Major Project (No.2021SHZDZX0103), Shanghai Science and Technology Innovation Action Plan (No. 22511104700), and National Natural Science Foundation of China (No. 62102095). Yanghua Xiao is also a member of Research Group of Computational and AI Communication at Institute for Global Communications and Integrated Media, Fudan University.

\clearpage

\bibliographystyle{ACM-Reference-Format}
\balance
\bibliography{main}

\end{document}